%% file: main.tex
\documentclass[10pt,twocolumn,letterpaper]{article}

\usepackage{cvpr}              
\usepackage{multirow}
\usepackage{tikz}
\usepackage{bbding}
\usepackage{amssymb}
\usepackage{amsmath}


\definecolor{cvprblue}{rgb}{0.21,0.49,0.74}
\usepackage[pagebackref,breaklinks,colorlinks,citecolor=cvprblue]{hyperref}

\def\paperID{6799} 
\def\confName{CVPR}
\def\confYear{2025}

\title{Is this Generated Person Existed in Real-world? Fine-grained Detecting and Calibrating Abnormal Human-body}

\author{Zeqing Wang\\
Sun Yat-sen University\\
Guangzhou, China\\
{\tt\small wangzq73@mail2.sysu.edu.cn}
\and
Qingyang Ma\\
Sun Yat-sen University\\
Guangzhou, China\\
{\tt\small maqy23@mail2.sysu.edu.cn}
\and
Wentao Wan\\
Sun Yat-sen University\\
Guangzhou, China\\
{\tt\small wanwt3@mail2.sysu.edu.cn}
\and
Haojie Li\\
South China \\University of Technology\\
Guangzhou, China\\
{\tt\small 12hjli4@gmail.com}
\and
Keze Wang\\
Sun Yat-sen University\\
Guangzhou, China\\
Peng Cheng Laboratory\\
Shenzhen, Guangdong, China\\
{\tt\small kezewang@gmail.com}
\and
Yonghong Tian\\
Peking University\\
Beijing, China\\
Peng Cheng Laboratory\\
Shenzhen, Guangdong, China\\
{\tt\small yhtianpku.edu.cn}
}

\begin{document}
\maketitle


\input{sec/0_abstract}

\input{sec/1_intro}

\input{sec/2_related_work}
\input{sec/3_task_datasets}

\input{sec/4_method}

\input{sec/5_exp}
\input{sec/7_conclusion}
{
    \small
    \bibliographystyle{ieeenat_fullname}
    \bibliography{main}
}

\input{sec/X_suppl}

\end{document}

%% file: sec/0_abstract.tex
\begin{abstract}
Recent improvements in visual synthesis have significantly enhanced the depiction of generated human photos, which are pivotal due to their wide applicability and demand. Nonetheless, the existing text-to-image or text-to-video models often generate low-quality human photos that might differ considerably from real-world body structures, referred to as ``abnormal human bodies''. Such abnormalities, typically deemed unacceptable, pose considerable challenges in the detection and repair of them within human photos. These challenges require precise abnormality recognition capabilities, which entail pinpointing both the location and the abnormality type. Intuitively, Visual Language Models (VLMs) that have obtained remarkable performance on various visual tasks are quite suitable for this task. However, their performance on abnormality detection in human photos is quite poor.
Hence, it is quite important to highlight this task for the research community. In this paper, we first introduce a simple yet challenging task, i.e., \textbf{F}ine-grained \textbf{H}uman-body \textbf{A}bnormality \textbf{D}etection \textbf{(FHAD)}, and construct two high-quality datasets for evaluation. Then, we propose a meticulous framework, named HumanCalibrator, which identifies and repairs abnormalities in human body structures while preserving the other content. Experiments indicate that our HumanCalibrator achieves high accuracy in abnormality detection and accomplishes an increase in visual comparisons while preserving the other visual content.

\end{abstract}

%% file: sec/1_intro.tex
\section{Introduction}
\label{sec:intro}
Visual content generation models have demonstrated the capacity to create highly realistic representations within human photos, which hold considerable importance across various downstream tasks such as virtual reality, augmented reality, and the entertainment industry. Recent developments in text-to-image~\cite{rombach2022high, ramesh2021zero, ramesh2022hierarchical, betker2023dalle3} and text-to-video models~\cite{podell2023sdxl, open_sora, yang2024cogvideox} have enhanced both the quality and the realistic of generated human photos. However, these models frequently struggle to accurately replicate human body structures as they exist in the real world, leading to human photos with abnormalities such as absent or redundant body parts. Compared to low-quality, this abnormality is more unacceptable because it is more noticeable and has a larger gap with the real-world human body structure.

Some methods try to solve this problem by adding additional constraints, such as Pose-ControlNet~\cite{zhang2023adding}, HumanSD~\cite{ju2023humansd}, and T2I-Adapter~\cite{mou2024t2i}. However, these methods are always hard to use due to their extra input or additional training. HumanRefiner~\cite{fang2024humanrefiner}, in another way, tackles the problem as a post-process method. It detects abnormalities in the generated human photos and then regenerates the whole content. However, this coarse-grained detection method can only detect which existing visual content is abnormal, ignoring the absent part. Furthermore, such a method can not preserve the background information of the original photo, which also limits its generalizability.

\begin{figure}[t]
  \centering
   \includegraphics[width=1\linewidth]{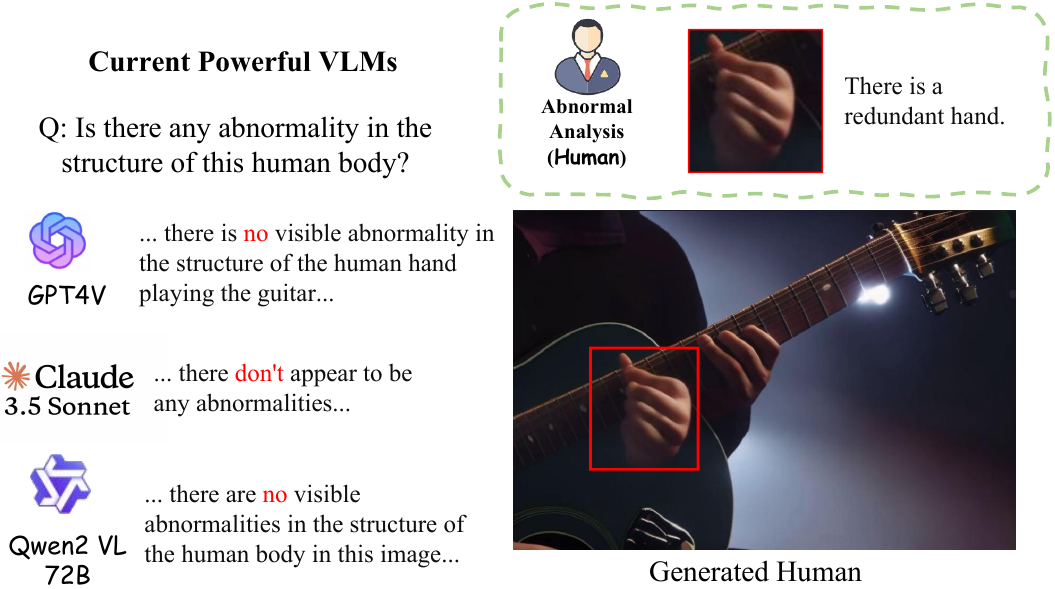}
   \vspace{-10pt}
   \caption{Fine-Grained Human-body Abnormality Detection (FHAD). Human body structures in AIGC often exhibit significant deviations from humans existing in the real world, making them easily recognizable as abnormal to human observers. However, current powerful VLMs, typically struggle with this abnormality perception despite excelling in various downstream perceptual tasks, which presents a challenge for fine-grained abnormality detection and motivates our research.}
   \label{fig:overview}
   \vspace{-10pt}
\end{figure}

To detect abnormalities, it is intuitive to use a Vision-Language Model (VLM), which processes strong perception and reasoning capabilities and has been applied to various downstream perception tasks~\cite{li2024llava_med,dai2024pa,prajapati2024evaluation}, as the backbone. However, when we test current powerful VLMs on our proposed Fine-grained Human-body Abnormality Detection (FHAD) task (Sec.~\ref{sec:task_define}), as shown in Figure~\ref{fig:overview}, their performance is surprisingly poor, though the task is simple for humans. Quantitative analysis across multiple models in Figure~\ref{fig:vlm_baseline} further demonstrates the lack of abnormality perception capabilities in existing VLMs. 

With the observation, we review the capabilities required to detect abnormalities. For absent body parts, detection relies on existing body parts and the correlation among them to infer the absent bodies. In contrast, for redundant body parts, this detection depends more on the overall perception of visual contents, as such redundant abnormalities may appear anywhere and are unrelated to the existing body parts.

With the aforementioned observation, we train a detection model that leverages the correlation of body parts to detect absent abnormalities. Furthermore, to tackle the redundant body parts, we employ a diffusion-based model that boasts strong capabilities in comprehending overall visual content. By integrating these approaches, we develop ``HumanCalibrator'', a fine-grained framework for the detection and repair of abnormalities in human body structure. Our HumanCalibrator precisely pinpoints the abnormalities within the human body structure and repairs the abnormal region while preserving other regions.

Our contributions can be summarized as follows: (i) To the best of our knowledge, we are the first to propose a simple yet challenging task, Fine-grained Human-body Abnormality Detection (FHAD) with two datasets across two main-stream domains 
(Sec.~\ref{sec:task_define}); (ii) Our extensive and comprehensive experiments demonstrate that it is a challenging task for the VLMs to understand abnormalities, despite being trained on large datasets and possessing strong ability in multiple downstream tasks (Sec.~\ref{subsec:exp_on_cur_vlms}). Then, leveraging the features of the human body structure, we propose a solution for absent and redundant body part abnormalities, which includes a detection model trained based on the correlation of human body structure (Sec.~\ref{subsec:absent_human_body_part}) and an approach focusing on the overall visual perception via diffusion-based model (Sec.~\ref{subsec:redundant_human_body_part}); (iii) We further propose an effective framework, i.e., HumanCalibrator, which includes detecting and repairing the abnormalities in the body structure while preserving the rest of the visual content (Sec.~\ref{subsec:p_a_c_framework}).

%% file: sec/2_related_work.tex
\section{Related Work} \label{sec:related_work}

\textbf{Vision-Language Model:} Leveraging extensive pre-training datasets and benefitting from Large Language Models (LLMs)~\cite{gpt3,llama,llama2,zeng2022glm,team2024gemma,jiang2023mistral,bai2023qwen,yang2024qwen2,bi2024deepseek,flant5} along with powerful Vision Encoders~\cite{blip2,vit,tong2024cambrian1,he2024bunny}, VLMs~\cite{blip2,liu2023llava,he2024bunny,tong2024cambrian1,dai2023instructblip,hu2024bliva,li2024mgm,xue2024xgen_blip3,chen2024far,radford2021learning} have achieved success in a range of visual tasks. These models excel at various perception and reasoning tasks~\cite{scienceQA, AOKVQA, gqa, ke2019reflective}, prompting research that fine-tunes VLMs for specific applications~\cite{zhong2024let,chen2024motionllm,Ren2023TimeChat,hanoona2023GLaMM}. However, due to these VLMs being mainly based on text-image alignment training strategies, our experiments demonstrate that even the most powerful VLMs(such as OpenAI's GPT4o) still fall short of detection abnormalities. 

\textbf{Detection in AIGC:} Detection within AIGC comprises various tasks; some initiatives aim to discern AIGC products~\cite{gramnet,f3net,univfd,luo2024lare,chang2024whatmatters}. Recently,~\cite{fang2024humanrefiner} tried to fix the abnormalities in the existing body parts, although it is limited to providing only coarse-grained results, leading to inconsistencies with the original visuals. In contrast, we introduce a novel task in abnormality detection,i.e., the FHAD. To the best of our knowledge, this is the first attempt at such Fine-Grained abnormality detection in AIGC products.

\begin{figure}[t]
  \centering
   \includegraphics[width=1\linewidth]{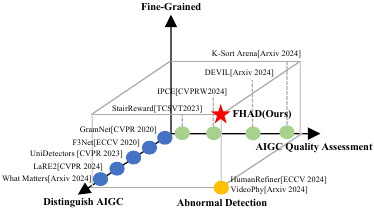}
   \vspace{-10pt}
   \caption{Fine-Grained Human-body Abnormality Detection \textbf{(FHAD)} (\textcolor{red}{$\bigstar$}) is a novel task. It is distinct from AIGC detection and the assessment of AIGC product quality, as its objective is to identify the abnormality of content generated by AIGC methods in relation to the real world. Additionally, detection at a fine-grained level necessitates methods capable of providing detailed information about the abnormalities and their locations.}
   \label{fig:compare_target}
   \vspace{-10pt}
\end{figure}

\textbf{Evaluation in AIGC:} Other methodologies interpret detection as a quality assessment exercise, evaluating AIGC products from diverse angles~\cite{li2024ksort_arena,ipce,stairreward}. For instance, VideoPhy~\cite{bansal2024videophy} assesses videos based on physical common sense, DEVIL~\cite{devil} focuses on dynamic quality, and some studies assess the overall quality of AIGC videos with text-image alignment and video characteristics. Our proposed method, distinct from these, strives to ascertain whether the generated human body structure could occur in the real world. Figure~\ref{fig:compare_target} presents a comparative analysis of our task objective across three aspects: (i) distinguish AIGC-produced content by identifying AIGC products involves verifying whether the content originated from AIGC models; (ii) assess the AIGC content's quality; (iii) detect the abnormality of AIGC content by spotting variances between generated outputs and real-world objects. 

\textbf{Visual Content Generation:} The realm of visual content generation has undergone considerable evolution. Initial endeavors, such as GAN-based architectures~\cite{goodfellow2014generative, mansimov2015generating, zhang2017stackgan, xu2018attngan} and autoregressive methodologies~\cite{ramesh2021zero, yu2022scaling, ding2021cogview}, provide foundational breakthroughs but encountered issues like low resolution and stability concerns. Then, with the advent of diffusion models including text-to-image~\cite {nichol2021glide, saharia2022photorealistic, rombach2022high, ramesh2022hierarchical} and text-to-video~\cite{singer2022make, ho2022imagen, podell2023sdxl, open_sora, yang2024cogvideox} , the quality of visual content has been further developed. However, the content generated by these models is often quite different from the real world, especially in the structure of the human body which limits their widespread downstream use.

%% file: sec/3_task_datasets.tex
\section{Task Definition}
\label{sec:task_define}
\textbf{Fine-grained Human-body Abnormality Detection (FHAD):} The goal of \textbf{FHAD} is to identify the differences in body structure from real-world humans in any given visual content that includes human photos. To achieve FHAD, the method needs to output the following two parts: (1) the semantic flag of abnormality $S^a \subseteq \textbf{A}$, that is, what type of body part abnormality $a$ exists. (2) the location of the abnormality $a$, output in the form of a bounding box $B^a$. For the input visual content within body part abnormalities $X$ and the pre-defined abnormalities set $\textbf{A}$, we consider any detection method as $M^d$, the task can be formatted as: 
\begin{equation}
[B^a, S^a ] = M^d(X).
\end{equation}
As shown in Figure~\ref{fig:overview}, for a given human photo, the method needs to detect the redundant hand in the \textcolor{red}{red} bounding box. 

\textbf{Human-body Abnormality Define:}  After reviewing a large number of generated human photos, we conduct an analysis of body part abnormalities and identified 12 distinct body part abnormalities which often create a significant gap between the real-world human body structure. It contains two types, the absent and redundant body parts. For the class of body parts, we identify the following body parts, i.e., head, ear, hand, arm, leg, and foot for $\textbf{P}$. Our following experiments are all conducted based on predefined abnormalities (\textit{e.g., absent hand, redundant hand...  $ \subseteq\textbf{A}$ }). 

\begin{figure}[t]
  \centering
   \includegraphics[width=1\linewidth]{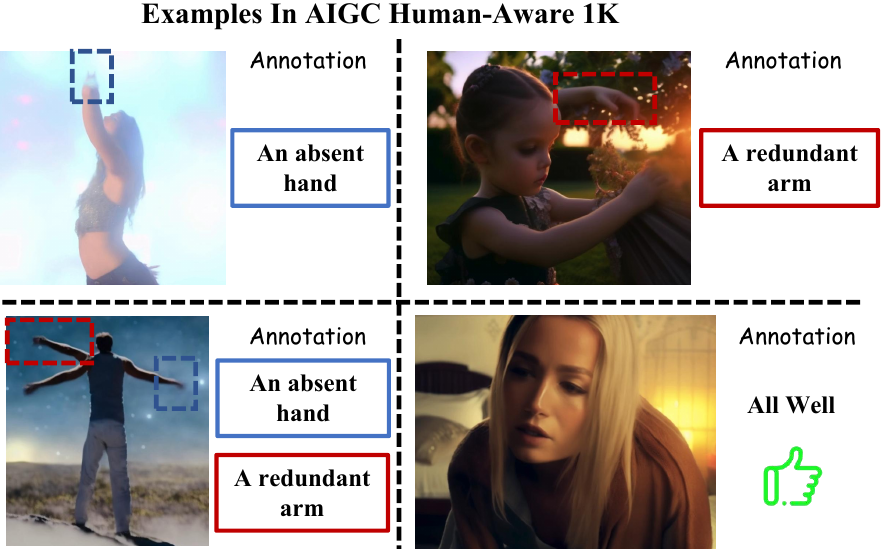}
   \vspace{-10pt}
   \caption{Examples in AIGC Human-Aware 1K. We manually annotate the abnormalities in frames from generated AIGC videos. Since the location of the abnormalities is ambiguous, we do not annotate the bounding box. Instead, we evaluate the accuracy of the bounding box location by assessing the repair quality.}
   \label{fig:aigc_1k_datasample}
   \vspace{-10pt}
\end{figure}

\textbf{FHAD Dataset:}
\label{subsec:abnormal_human_dataset}
We propose two datasets for the proposed FHAD task, i.e., the COCO Human-Aware Val and the AIGC Human-Aware 1K. For \textbf{COCO Human-Aware Val}, we adopt an automated dataset construction method to construct images with pre-defined absent abnormalities from the full COCO Val split. We replace one body part with the background to make an absent body part abnormality. For \textbf{AIGC Human-Aware 1K}, it is a cross-domain, meticulously hand-labeled dataset with \textbf{\textit{inherent}} body part abnormality. In order to enhance the generalization of the AIGC Human-Aware 1K. We build up this dataset based on the large generated video dataset VidProM~\cite{wang2024vidprom}. To ensure the basic quality of the generated videos, we randomly choose the videos from the PIKA split with human photos. Then, we manually annotate 1K samples with the same body part classification in frame level, some cases are shown in Figure~\ref{fig:aigc_1k_datasample}. Please note that compared to the COCO Human-Aware Val, AIGC Human-Aware 1K is a more challenging dataset for VLM as it comes from the AIGC domain and contains both absent and redundant body parts, which is the main evaluation dataset for our task. We provide a detailed dataset construction process in Appendix~\ref{sec:data_annotation}, and we \textbf{highly recommend} reading this section, as it will be of great help in understanding our task. We also provide cases in COCO Human-Aware Val in Appendix~\ref{sec:coco_val_case}.

%% file: sec/4_method.tex
\section{Methodology}
\label{sec:method}

\subsection{Solution for Absent Body Part Detection}
\label{subsec:absent_human_body_part}
Within a detection pipeline, suppose the predefined human body classes of the whole body part as $\textbf{P}$ with the corresponding bounding box $\textbf{B}$ for the given visual content within human photo $X$. The existing body parts are represented as a set $\textbf{P}^e \subseteq \textbf{P}$ with their corresponding bounding boxes $\textbf{B}^e \subseteq \textbf{B}$. The evaluated method $M^a$ should present the absent body part $p^a$ and the absent area $b^a$. Note that, $X$ also includes context for situations, e.g., obstructions, to avoid misjudgments of absent body parts. It can be represented as follows:
\begin{small} 
\begin{equation}
    \{\left \langle p^a_i, b^a_i\right \rangle\}_{i=0}^n = M^a(\textbf{P}^e, \textbf{B}^e; X),
\label{eq:absent_method}
\end{equation}
\end{small}
where $n$ denotes the number of absent body parts. 

For this task, the VLM is an intuitive choice since it is trained on a vast dataset and exhibits strong capabilities across a wide range of downstream tasks. However, the results demonstrate that (Sec.~\ref{subsec:exp_on_cur_vlms}), though VLMs have been trained with a large quantity of normal data like ours, they still lack awareness of the abnormality on COCO Human-Aware Val which is based on real-world images and only contain absent abnormalities. We discuss this phenomenon in Appendix~\ref{sec:why_lack}. To solve this problem, the simplest and most intuitive method is to manually annotate a large training dataset on real AIGC data like~\cite{fang2024humanrefiner,bansal2024videophy}. However, for the tasks we propose, extensive manual annotation is unrealistic and extremely costly (we perform the detailed annotation process in Appendix~\ref{sec:data_annotation}). Furthermore, data annotated based on a specific generative model will inevitably contain certain biases. Therefore, our goal shifted towards automating the construction of these training data from a real-world dataset, which is also more aligned with our target.

\begin{figure}[t]
  \centering
   \includegraphics[width=1\linewidth]{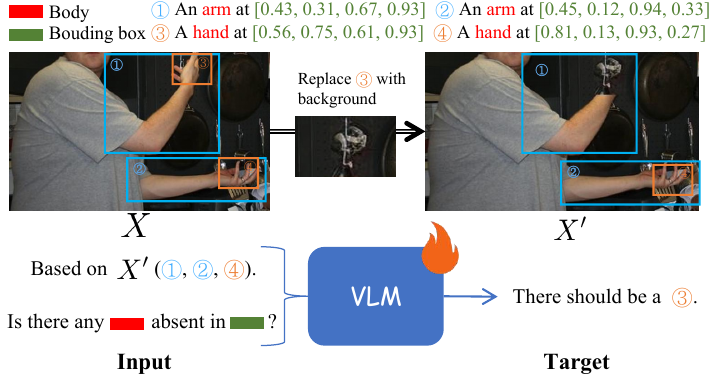}
   \vspace{-10pt}
   \caption{Absent Human-body Detector (AHD) training strategy. In the real world, many objects within the visual content are interconnected, meaning that based on the other objects, one can infer the presence of certain objects in specific locations. Our proposed training strategy leverages the correlation between body parts to facilitate this training process.}
   \label{fig:training_stragy}
   \vspace{-10pt}
\end{figure}

Inspired by the process of humans detecting absent abnormalities, we propose the following \textbf{body part correlation} training strategy. For the given visual content $X$ with a normal human, we first ground all its body parts $\{\left \langle p_i, b_i\right \rangle\}_{i=0}^n$, where $n$ is the number of grounded body parts, based on our predefined set $\textbf{P}$. For each body part representation $\left \langle p_i, b_i\right \rangle$, we apply a mask operation to obtain:
\begin{small}
\begin{equation}
X_i = mask(X, \left \langle p_i, b_i\right \rangle),
\end{equation}
\end{small}
where $mask$ replaces the area $b_i$ with the background of $X$. For each masked image, we maintain:
\begin{small}
\begin{equation}
\{(X_i, \left \langle p_i^a, b_i^a\right \rangle)\}_{i=0}^n = \{(X_i, \left \langle p_i, b_i\right \rangle)\}_{i=0}^n,
\end{equation}
\end{small}
to get the absent body part training sample $(X_i, \left \langle p_i^a, b_i^a\right \rangle)$. Similar to the current training objective of VLMs and LLMs, for each masked image $X_i$ and its corresponding absent body part representation $\left \langle p_i^a, b_i^a\right \rangle$, we optimize the following auto-regressive objective:
\begin{small}
\begin{equation}
\begin{split}
        &p(\left \langle p^a_i, b^a_i\right \rangle|X_i, I_a) = \\ &\prod_{j=1}^{L}p_\theta(x_j|X_i, I_a, _{<j}, \left \langle p^a_i, b^a_i\right \rangle, _{<j}),
\end{split}
\label{eq:vlm_trianing}
\end{equation}
\end{small}
where $\theta$ represents the trainable parameters of the VLM, $L$ is the length of the concatenated instruction $I_a$ and the perception and position of the current absent body part $\left \langle p^a_i, b^a_i\right \rangle$. This training process sequentially replaces body parts in normal images with the background, as shown in Figure~\ref{fig:training_stragy}, allowing the VLM to learn the correlation between absent and existing body parts in terms of position and class based on the existing body parts.

Based on this training method, we develop a VLM named Absent Human-body Detector (\textbf{AHD}), represented as $D$, which is capable of detecting absent abnormalities of the body part. The AHD can infer from a given human photo whether there is any absent body, as well as identify the location and content of these absent body parts.

\begin{figure*}[!t]
  \centering
   \includegraphics[width=1\linewidth]{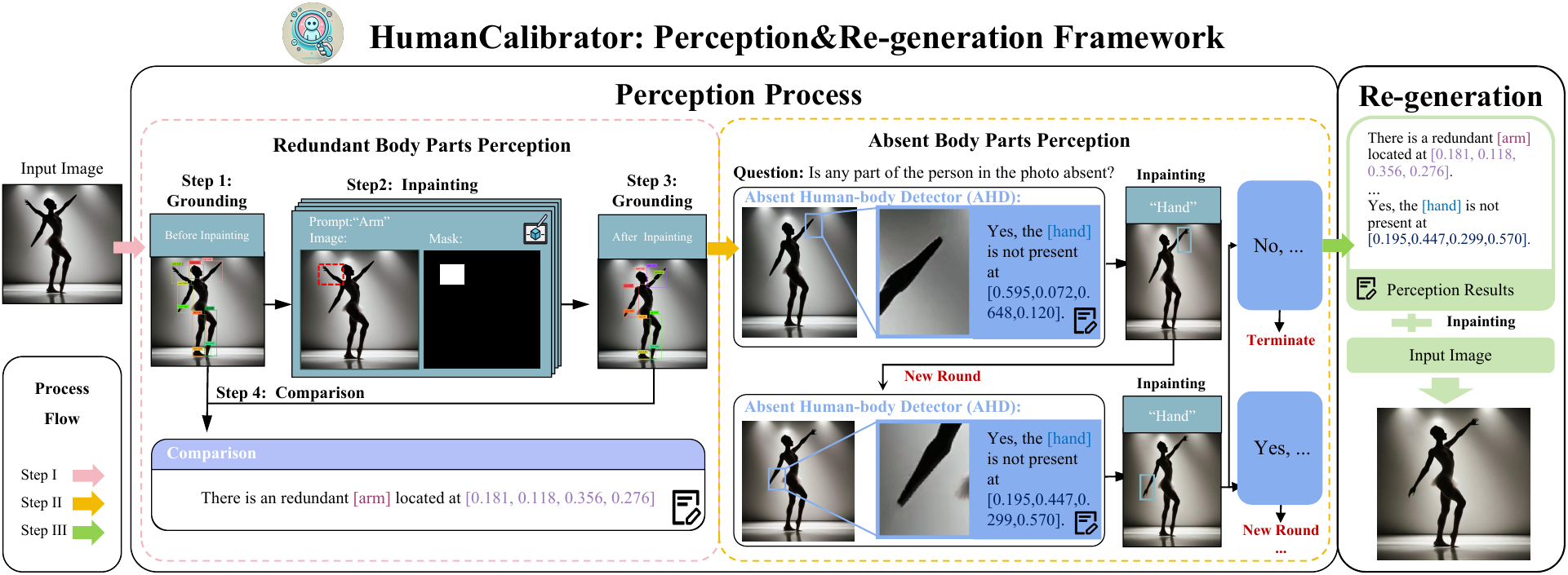}
   \vspace{-15pt}
   \caption{The illustration of our HumanCalibrator. The HumanCalibrator consists of two parts: perception and regeneration. In the perception stage, HumanCalibrator initially uses an inpainting model to re-generate various bodies based on its overall understanding of human body structure, determining the redundant bodies by comparing semantic differences before and after inpainting. Subsequently, relying on our Absent Human-body Detector (AHD) to assess the perception of absent abnormalities, our HumanCalibrator employs a cyclical strategy to identify absent bodies via AHD. Finally, by the results of the perception stage into the inpainting model as prompts, our HumanCalibrator can repair the detected abnormalities while preserving the visual content of the remaining areas.}
   \label{fig:overview_pipline}
   \vspace{-10pt}
\end{figure*}

\subsection{Solution For Redundant Body Part Detection}
\label{subsec:redundant_human_body_part}
Compared to absent body part abnormality, the situation of redundant body part abnormality is more diverse, which is reflected in the following two parts: (1) The position of the redundant body parts, which can appear in any area of the given photo. (2) The number of redundant body parts, which can be arbitrary. These two factors make the judgment no longer dependent on the existing body parts $\textbf{P}^e$. For each redundant body parts $p^r$ with its corresponding area $b^r$, for given visual content $X$ and the detection method $M^r$ can be formalized as:
\begin{small}
\begin{equation}
\label{eq:task_for_redundant}
    \{\left \langle p^r_i, b^r_i\right \rangle\}_{i=0}^n = M^r(X),
\end{equation}
\end{small}
where $n$ is the number of redundant body parts. In contrast with Eq.~\ref{eq:absent_method}, Eq.~\ref{eq:task_for_redundant} indicates that addressing redundant bodies relies more on global visual information $X$ rather than the existing body parts $\textbf{P}^e$ within $X$. Based on this, we utilize a diffusion-based inpainting model $R$ with strong contextual understanding capabilities, combined with the grounding model $G$, to detect redundant body parts. In detail, for the given $X$, the model $G$ can ground all body parts $p^g_i \in \textbf{P}$ with their locations $b^g_i$, as:
\begin{small}
\begin{equation}
\label{eq:get_the_masks}
    \{\left \langle p^g_i, b^g_i\right \rangle\}_{i=0}^n = G(X, \textbf{P}),
\end{equation}
\end{small}
where $n$ is the number of body parts in the given $X$. Then for each $\left \langle p^g_i, b^g_i\right \rangle$, we use the model $R$ to re-generate the content of $b^g_i$ with the text-condition $p^g_i$, represented as $p^R$. Trained on large datasets of normal body structure without any redundant bodies, given the $\left \langle p^r, b^r\right \rangle$ as mask location and text-condition for $R$, the semantic $p^R$ of model $R$'s output content at $b^r$ tend to exihibit a significant difference from the original $p^r$.
For example, when a mask is applied to the location of a redundant arm, R is more likely to generate background in that area rather than the redundant arm itself. 
To determine if the original body part $b_i^g$ is indeed the redundant body $b^r$, we compare the corresponding semantics $p_i^g$  to $p^R$. If a significant semantic difference is detected (with the assistance of G), it indicates that the body part $\left \langle p_i^g, b_i^g \right \rangle$ is redundant.

This process can be formalized as:
\begin{small}
\begin{equation}
\label{eq:eq_for_redundant}
    \{\left \langle p^r_k, b^r_k\right \rangle\}_{k=0}^j = \{G(R(p^g_i, b^g_i), p^g_i)\}_{i=0}^n < \tau,
\end{equation}
\end{small}
where $j$ is the number of the detected redundant body parts and $\tau$ is the grounding threshold.

\subsection{HumanCalibrator}
\label{subsec:p_a_c_framework}
By leveraging the proposed Absent Human-body Detector $D$ with absent body part abnormality perception in Sec~\ref{subsec:absent_human_body_part}, and our proposed method for handling redundant body parts in Sec~\ref{subsec:redundant_human_body_part} we develop a comprehensive framework named HumanCalibrator, for the proposed Fine-Grained Human-body Abnormality Detection (FHAD) task. Furthermore, leveraging the ability of fine-grained detection, our HumanCalibrator can repair abnormalities of body parts while preserving other visual content unchanged. The details of HumanCalibrator are shown in Figure~\ref{fig:overview_pipline}. In detail, our proposed framework can be divided into the following three steps:

\begin{itemize}
    \item \textbf{Step I:} Detect redundant body parts in the given visual content $X$. The first step is to obtain the set of redundant body parts in $X$ via Eq.~\ref{eq:get_the_masks}, Eq.~\ref{eq:eq_for_redundant},i.e., $\{\left \langle p^r_k, b^r_k\right \rangle\}_{k=0}^j$, where $k$ is the number of redundant body parts. Identifying redundant body parts first serves two purposes. (1) Provide a better image base for addressing absent body part abnormalities. (2) Self-refine during the process of addressing absent human body abnormalities. If it is found that the detected absenting body part is the same as the previously identified redundant bodies, it can be considered that this is a wrongly resolved redundant body, thereby improving the accuracy of the entire framework.
    \item \textbf{Step II:} Detect cyclically absent body parts in the given visual content $X$. We use  $D$ to detect the absent body part in the current $X$ and obtain the detection result  $\langle p^a, b^a \rangle$. Subsequently, we utilize the inpainting prompt template  $T$, which is predefined according to  $\textbf{P}$, to obtain the corresponding repair prompt  $T(p^a)$. By combining $p^a$ and $b^a$, we use the inpainting model $R$ to obtain a new image $X'$. The resulting $X'$  is then used as the image for the next detection cycle. This process continues until $D$ determines that there are no new $\langle p^a, b^a \rangle$ in the current $X'$, which can be formalized as:
    \begin{small}
    \begin{equation}
    X_{t+1} = \begin{cases}
    R(X_t, T(p^a), b^a), & \text{if}~ \langle p^a, b^a \rangle = D(X_t) \neq \emptyset \\
    X_t, & \text{otherwise}.
    \end{cases}
    \end{equation}
    \end{small}
    Through this loop, we can obtain all the absent bodies in the image, denoted as $\{\left \langle p^a_i, b^a_i\right \rangle\}_{i=0}^n$, where $n$ is the number of detected absent body parts.
    \item \textbf{Step III:} Repair the abnormality detected above. After detecting the redundant and absent body parts separately, we can repair the detected abnormalities through the Inpainting model $R$. We start with the original image $X$ and repair these abnormalities one by one. We also use the pre-defined inpainting prompt template $T$ to repair different types of abnormalities, which can be formalized as:
    \begin{small} 
    \begin{equation}
    \begin{split}
    &X_0 = X \\
    &X_{t+1} = R(X_t, T(p_t), b_t),~\text{for}~t = 0,\dots,j+n \\
    &\text{where}\langle p_t, b_t \rangle = \begin{cases}
    \langle p^r_t, b^r_t \rangle, & \text{if}~0 \leq t \leq j \\
    \langle p^a_{t-j}, b^a_{t-j} \rangle, & \text{if}~j < t \leq j+n,
    \end{cases}
    \end{split}
    \end{equation}
    \end{small}
    where $j$ and $n$ represent the number of redundant and absent body parts, respectively. The final $X_{t+j}$ is the image after all abnormalities have been repaired.

\end{itemize}

%% file: sec/5_exp.tex
\section{Experiment}
\label{sec:exp}

\begin{table*}[!t]
\centering
\resizebox{2\columnwidth}{!}{
  
    \begin{tabular}{cccccccccccccccc}
    \toprule
    Type  & \multicolumn{15}{c}{Human Body} \\
    \midrule
    \multirow{7}[6]{*}{Absent} & \multirow{2}[4]{*}{Method} & \multicolumn{2}{c}{hand} & \multicolumn{2}{c}{leg} & \multicolumn{2}{c}{ear} & \multicolumn{2}{c}{foot} & \multicolumn{2}{c}{arm} & \multicolumn{2}{c}{head} & \multicolumn{2}{c}{Avg} \\
\cmidrule{3-16}          &       & Acc $\uparrow$   & FDR $\downarrow$   & Acc $\uparrow$   & FDR $\downarrow$  & Acc $\uparrow$   & FDR $\downarrow$  & Acc $\uparrow$   & FDR $\downarrow$  & Acc $\uparrow$   & FDR $\downarrow$  & Acc $\uparrow$   & FDR $\downarrow$  & Acc $\uparrow$   & FDR $\downarrow$\\
\cmidrule{2-16}          
          & LLaVA-34B & 0.42\% & \textbf{0.92\%} & 15.00\% & \textbf{2.14\%} & 0.47\% & \textbf{0.00\%} & 0.00\% & \textbf{--} & 6.90\% & 5.26\% & 36.00\% & 5.74\% & 9.80\% & 2.34\% \\
          & InternVL2-26B & 2.95\% & 2.75\% & 10.00\% & 5.51\% & 5.69\% & 2.41\% & 3.03\% & 2.46\% & 42.53\% & 35.71\% & 52.00\% & 8.31\% & 19.37\% & 9.53\% \\
          & GPT-4o & 8.02\% & 2.49\% & 0.00\% & \textbf{--} & 0.47\% & 0.13\% & 10.61\% & \textbf{0.00\%} & 12.64\% & \textbf{3.83\%} & 20.00\% & \textbf{0.62\%} & 8.62\% & \textbf{1.38\%} \\
          & CLIP-Large-14 & 14.35\% & 6.55\% & 15.00\% & 4.59\% & 17.06\% & 16.22\% & 19.70\% & 2.68\% & 44.83\% & 16.10\% & 44.00\% & 5.74\% & 25.82\% & 8.65\% \\
          & HumanCalibrator(Ours) & \textbf{79.75\%} & 12.19\% & \textbf{75.00\%} & 6.22\% & \textbf{79.15\%} & 10.14\% & \textbf{90.91\%} & 4.39\% & \textbf{79.31\%} & 12.92\% & \textbf{80.00\%} & 5.03\% & \textbf{80.69\%} & 8.48\% \\
    \midrule
    \multirow{7}[6]{*}{Redundant} & \multirow{2}[4]{*}{Method} & \multicolumn{2}{c}{hand} & \multicolumn{2}{c}{leg} & \multicolumn{2}{c}{ear} & \multicolumn{2}{c}{foot} & \multicolumn{2}{c}{arm} & \multicolumn{2}{c}{head} & \multicolumn{2}{c}{Avg} \\
\cmidrule{3-16}          &       & Acc $\uparrow$  & FDR $\downarrow$  & Acc $\uparrow$  & FDR $\downarrow$  & Acc $\uparrow$  & FDR $\downarrow$  & Acc $\uparrow$  & FDR $\downarrow$  & Acc $\uparrow$  & FDR $\downarrow$  & Acc $\uparrow$  & FDR $\downarrow$  & Acc $\uparrow$  & FDR $\downarrow$\\
\cmidrule{2-16}          
          & LLaVA-34B & 3.16\% & \textbf{0.44\%} & \textbf{33.33\%} & 2.33\% & 0.00\% & \textbf{--} & 0.00\% & \textbf{--} & 48.57\% & 18.76\% & 14.29\% & 1.91\% & 16.56\% & 3.97\% \\
          & InternVL2-26B & 2.11\% & 1.10\% & 0.00\% & \textbf{--} & 16.67\% & \textbf{0.60}\% & 0.00\% & \textbf{--} & 20.00\% & 5.70\% & \textbf{57.14\%} & 0.70\% & 15.99\% & 1.64\% \\
          & GPT-4o & 7.37\% & 0.88\% & 0.00\% & \textbf{--} & 0.00\% & \textbf{--} & 1.71\% & \textbf{0.20\%} & 25.71\% & 3.83\% & 14.29\% & \textbf{0.60\%} & 8.18\% & \textbf{1.00\%} \\
          & CLIP-Large-14 & 21.05\% & 6.08\% & 25.00\% & 2.94\% & 33.33\% & 15.90\% & 0.00\% & \textbf{--} & \textbf{57.14\%} & 17.41\% & 0.00\% & \textbf{--} & 22.75\% & 7.96\% \\
          & HumanCalibrator(Ours) & \textbf{65.26\%} & 3.87\% & \textbf{33.33\%} & \textbf{0.81\%} & \textbf{83.33\%} & 2.31\% & \textbf{66.67\%} & 0.70\% & 45.71\% & \textbf{2.49\%} & \textbf{57.14\%} & 5.03\% & \textbf{58.57\%} & 2.54\% \\
    \bottomrule
    \end{tabular}%
    }
    \vspace{-5pt}
    \caption{Detection accuracy of body part abnormalities. We evaluate performance via acc and false discovery rate (FDR). The \textbf{--} implies that when the acc is zero, the FDR loses its statistical significance. The test is performed on open-source VLM (LLaVA-34B and InternVL2-26B), closed-source VLM (GPT4o), CLIP, and our HumanCalibrator. Compared to the best-performing existing VLM, our framework significantly improves accuracy across almost all abnormality types with low FDR. We provide a detailed evaluation process for the baseline and an analysis of the reasons for their poor performance in the Appendix~\ref{sec:detial_baselines}.}
    \label{tab:main_results}%
    \vspace{-5pt}
\end{table*}%

\begin{figure}[!t]
  \centering
   \includegraphics[width=1\linewidth]{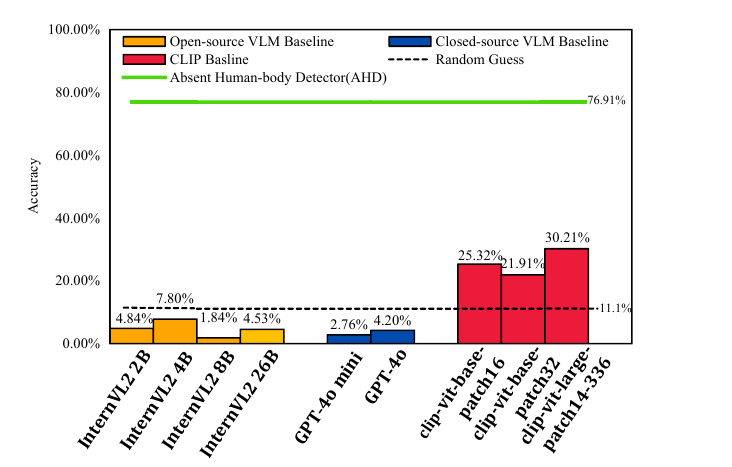}
   \vspace{-15pt}
   \caption{Comparison of accuracy in COCO Human-aware Val. We evaluate \textcolor{orange}{Open-source} VLMs (InternVL2) and \textcolor{blue}{Closed-source}  VLMs (GPT-4o and GPT-4o mini), \textcolor{red}{CLIP} and our \textcolor{green}{Absent Human-body Detector} (\textbf{AHD}). The powerful VLMs demonstrate significant limitations in perceiving abnormalities in body parts, with results often comparable to random guessing, even though the COCO Human-aware Val is built upon real-world images that are similar to their training data set.}
   \label{fig:vlm_baseline}
   \vspace{-10pt}
\end{figure}

\begin{table}[!t]
\small
  \centering
    \begin{tabular}{lrr}
    \toprule
    Metric & \multicolumn{1}{c}{Original} & \multicolumn{1}{c}{Ours} \\
    \midrule
    Human Concept Score $\uparrow$ & 22.59 & \textbf{22.77} \\
    CLIP Score $\uparrow$ & 41.87 & \textbf{41.97} \\
    Human CLIP Score $\uparrow$ & 26.36 & \textbf{26.42} \\
    \midrule
    Metric & \multicolumn{1}{c}{Pose Condition} & \multicolumn{1}{c}{Ours} \\
    \midrule
    FID $\downarrow$ & 98.86 & \textbf{16.55} \\
    Latent Consistency $\uparrow$ & 0.668 & \textbf{0.964} \\
    \bottomrule
    \end{tabular}%
    \vspace{-5pt}
    \caption{Repair quality and visual consistency evaluation. Repair quality is assessed via: (1) Human Concept Score, measuring the similarity between the person in the visual content with a real-world human. (2) CLIP Score, and (3) Human CLIP Score, focuses on the prompt describing the person, that is more relevant to our target. The repaired images show improvements over the original ones across these metrics. The modest improvement in the metrics is due to our effectiveness in preserving the remaining visual content. For visual consistency, we compare our method to a pose-conditioned method~\cite{fang2024humanrefiner} via FID and Latent Consistency showing that our approach maintains strong visual consistency.}
    \label{tab:human_and_consistence}%
    \vspace{-5pt}
\end{table}%

\textbf{Experiments on Exploring  Current VLMs.}
\label{subsec:exp_on_cur_vlms}
To objectively verify the ability of the existing VLM to detect the abnormalities of body structure, we employ the automatically generated COCO Human-Aware Val to test the VLMs. It is important to note that the COCO Human-Aware Val is automatically produced based on the entire COCO Val Split.

We compared different vision-language models (VLMs), including the state-of-the-art \textcolor{orange}{Open-source VLM} InternVL2 and \textcolor{blue}{Closed-source VLM} GPT4o and GPT4o-mini, along with contrastive learning-based models such as \textcolor{red}{CLIP}, as presented in Figure~\ref{fig:vlm_baseline}. The results demonstrate that despite their strong performance in many visual tasks, these models struggle with abnormal perceptions of body parts, displaying accuracies close to random guesses. Interestingly, even though humans and these VLMs are exposed to similar volumes of normal data, abnormality detection remains markedly easier for human cognition. The intricacies of our comparative analysis are further discussed in Appendix ~\ref{sec:why_lack}, and the specifics of the baseline models we test are detailed in Appendix~\ref{sec:detial_baselines}.

\begin{figure*}[!t]
  \centering
   \includegraphics[width=1\linewidth]{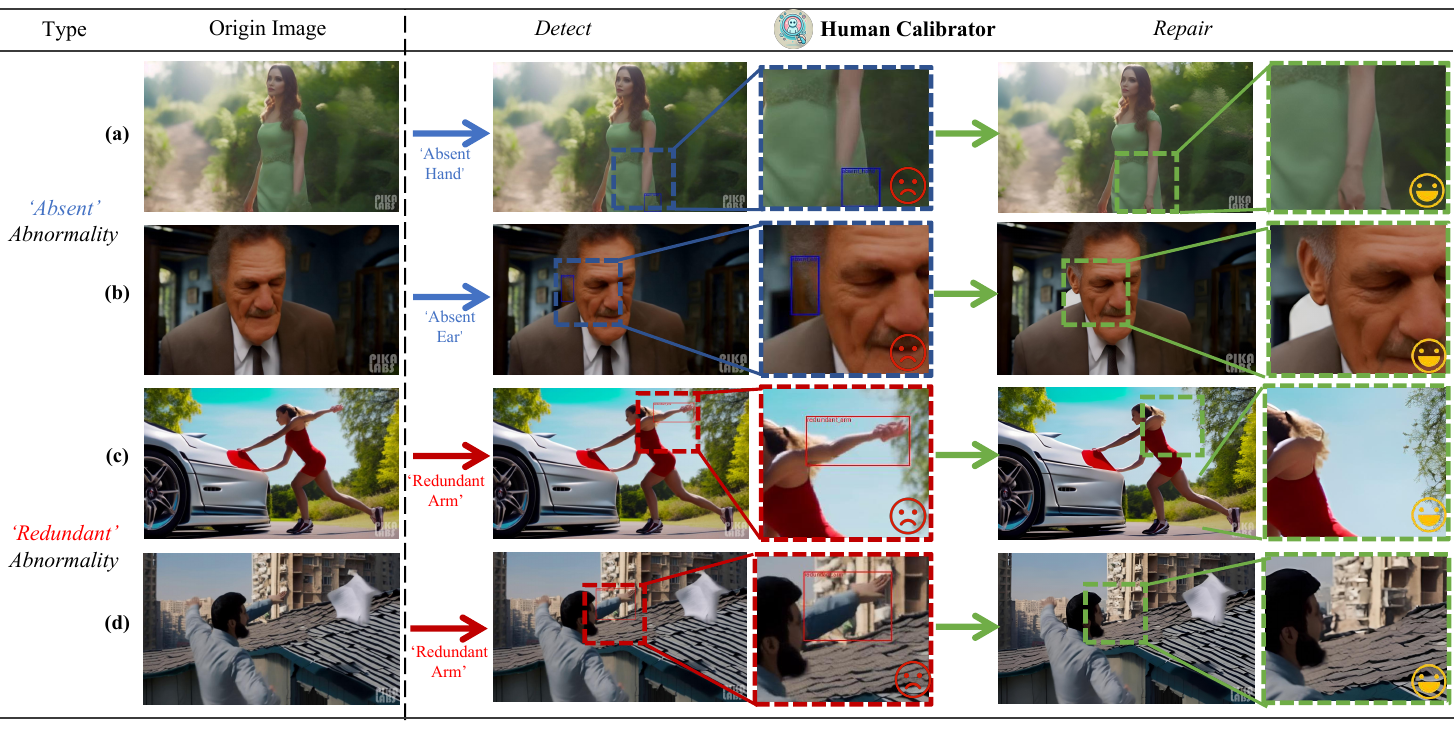}
   \vspace{-20pt}
   \caption{Case study of the image repair quality of HumanCalibrator. In (a), the HumanCalibrator detects the absent hand of the woman and repairs it with a hand in the correct pose and shape. In (b), an absent ear is identified, and the HumanCalibrator regenerates the ear without altering the man's face or expression. In (c), the redundant arm of the woman is detected and removed without affecting the background. In (d), the redundant arm of the person is corrected while preserving the rest of his body. Compared to the original images, our method achieves high-quality repair of the human body structure while preserving the remaining visual content.}
   \label{fig:image_case_study}
   \vspace{-10pt}
\end{figure*}

\textbf{Details of Human Calibrator:}
For the absent body part detector, we finetune the LLaVAv1.5 7B~\cite{liu2023llava} on COCO Train Split via Eq.~\ref{eq:vlm_trianing}. The format of the training data is similar to the COCO Human-Aware Val, some cases are shown in Appendix~\ref{sec:coco_val_case}. All training runs on 4 NVIDIA A800 GPUs. It takes around 30 hours to fine-tune 2 epochs with a learning rate of 2$\times10^{-5}$. We also test its ability on the COCO Human-Aware Val, the \textcolor{green}{results} are shown in Figure~\ref{fig:vlm_baseline}. For the other pretrained models used in our HumanCalibrator, we list more details in the Appendix~\ref{sec:detials_in_human_crafter}.

\begin{figure*}[!t]
  \centering
   \includegraphics[width=1\linewidth]{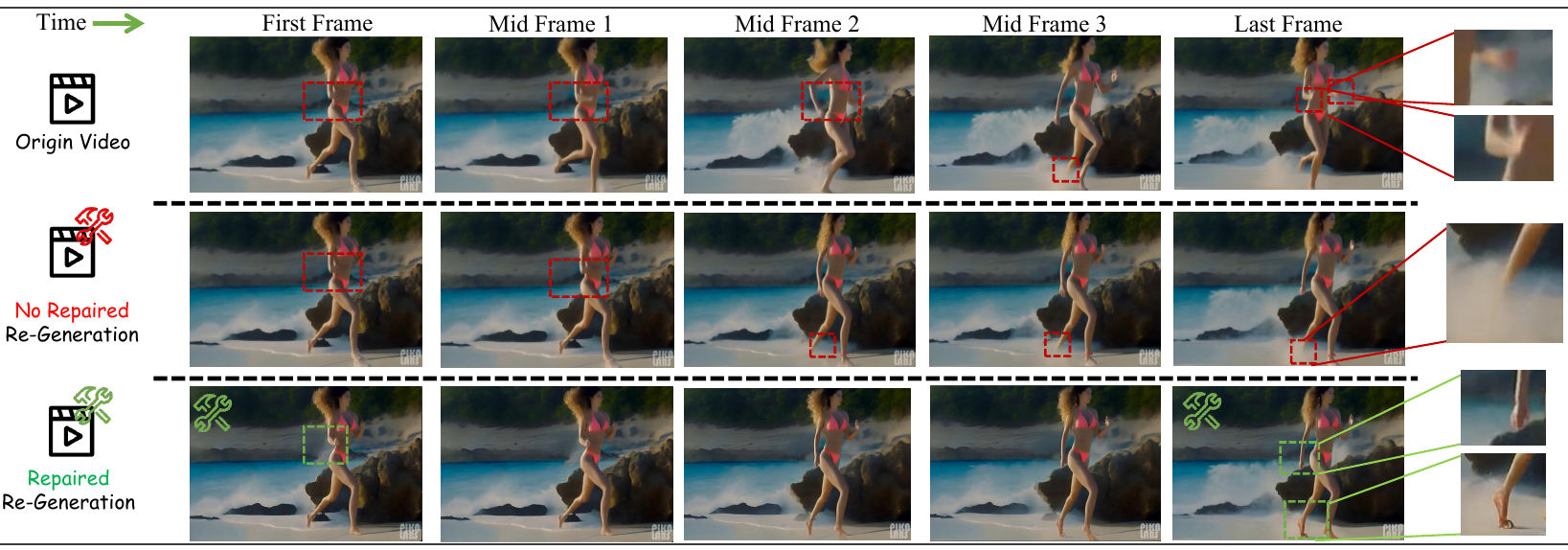}
   \vspace{-20pt}
  \caption{Case study of the video repair quality of HumanCalibrator. The first row shows frames from the original video, which contain noticeable abnormalities; the second row shows the video generated by a keyframe interpolation model using the original first and last frames, which preserves most of the visual information but still exhibits abnormalities; the third row shows the video regenerated with repaired first and last frame by HumanCalibrator, efficiently addressing the abnormalities while preserving the remaining visual information.}
   \label{fig:video_case_study}
   \vspace{-10pt}
\end{figure*}

\begin{figure}[!t]
  \centering
   \includegraphics[width=1\linewidth]{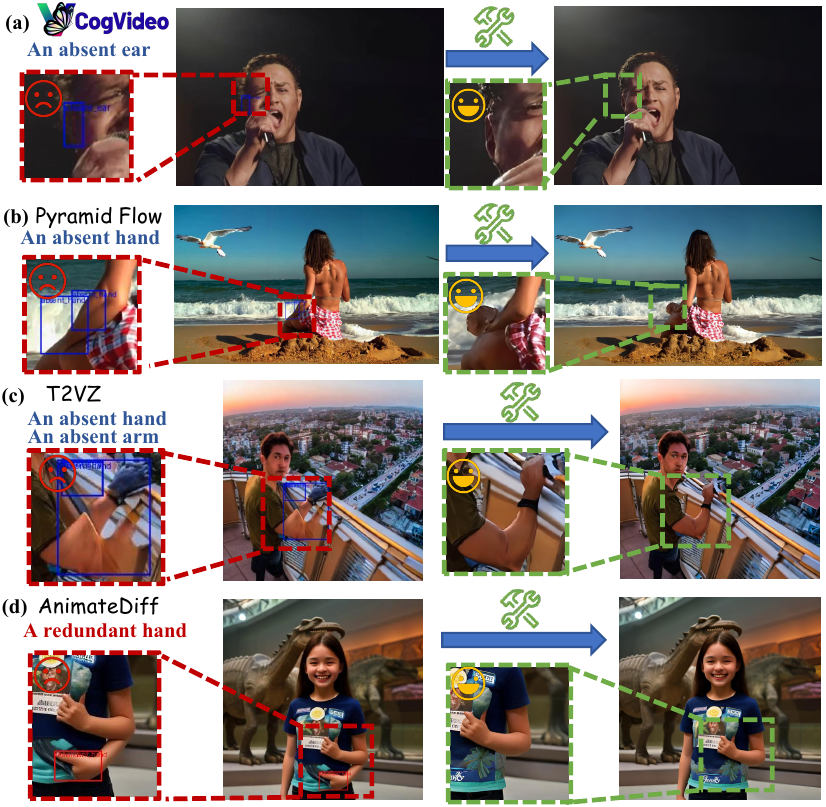}
   \vspace{-20pt}
  \caption{Examples of HumanCalibrator applied to frames generated by other SOTA Video generation models (frames in (a),(b),(c),(d) are produced by CogVideoX~\cite{yang2024cogvideox}, PyramidFlow~\cite{Lei_2023_CVPR}, T2VZ~\cite{text2video-zero} and AnimateDiff~\cite{guo2023animatediff}, respectively). Each model generates human photos with distinct abnormalities, which are subsequently addressed by our HumanCalibrator. In (a), the HumanCalibrator detects and repairs an absent ear on the singer.  In (b), an absent hand is identified, with HumanCalibrator identifying the positions and completing the restoration. In (c), the HumanCalibrator locates and repairs an absent hand and arm. In (d), the girl's redundant hand is detected and corrected.}
   \label{fig:other_models}
   \vspace{-10pt}
\end{figure}

\textbf{Body Part Abnormality Detection.}
We evaluate the accuracy of our Human Calibrator's perception ability of abnormalities on the AIGC Human-Aware 1K dataset. Similar to the assessment on the COCO Human-Aware Val, we also benchmark several recent powerful VLMs as our baseline, the results are shown in Table~\ref{tab:main_results}, and we provide the acc and false detection rate (FDR) for each specific category. Similar to the results on the COCO Human-Aware Val, existing VLMs have difficulty in accurately perceiving abnormalities. In the aspect of absent detection, the trained Absent Human Detector effectively detects the existing abnormalities while maintaining a low FDR. In terms of absences, even under training-free conditions, our method achieves better results compared to the baseline. We analyze these results in detail in Appendix~\ref{sec:detial_baselines}.

\subsection{Body Part Abnormality Repair}
\textbf{Metrics.} The goal of the task we proposed is to detect and locate the abnormality of body parts that make the human different from real-world humans, to comprehensively evaluate the effectiveness of our proposed HumanCalibrator in solving the task, we adopt six metrics for quantitative evaluation: (1) accuracy and FDR of abnormality detection, which calculate the detection results for each category of absent and redundant body parts. (2) Clip Score, the similarity between the image and the original prompt.
(3) Human CLIP Score, the similarity between the image and the prompt which only contains only descriptions related to the human. (4) Human Concept Score, the similarity to the concept of `human'. (5) FID, we use the origin image as the real image and the repaired image as the generated image to assess the extent to which our repair method preserves the original information. (6) Latent Consistency, consistency between the original and repaired image in the latent space.

\textbf{Repair Quality} provides insight into the overall performance from two aspects. (1) The accuracy of the detected bounding boxes, due to the worse repair results caused by inaccurate re-generation to the detected areas compared to the original image. (2) Whether our repair makes the person in the visual content more similar to a real-world human, which is the motivation of our proposed task. CLIP Score and Human CLIP Score are used to evaluate the quality of the repairs, with the Human CLIP Score focusing on the prompt describing the person, which is more closely related to our task. The Human Concept Score measures the distance between the human in the image and a `human' in the real world. As shown in Table~\ref{tab:human_and_consistence}. Compared to the original visual content, the repaired images show a certain degree of improvement in various metrics. The extent of our improvement is not significant, which is due to our good maintenance of consistency outside the abnormal parts of the image. To evaluate this consistency, we further measured the metrics on visual consistency.

\textbf{Visual Consistency.} A key advantage of our proposed fine-grained anomaly detection is its ability to repair only the abnormalities while maintaining the consistency of other information. We compare our method with the pose-condition-based abnormality repair method~\cite{fang2024humanrefiner} in FID and Latent space, as shown in Table~\ref{tab:human_and_consistence}. It can be observed that our HumanCalibrator maintains good visual consistency at both metrics. The details of the evaluation are shown in Appendix~\ref{sec:detial_of_metrics}.

\textbf{Case Study.} We present three levels of case studies: image-level, video-level, and generalization-level. (1) Image Case Study: As shown in Figure~\ref{fig:image_case_study}, owing to the fine-grained abnormality detection of our HumanCalibrator, it repairs the abnormalities in the human body structure within images while preserving other normal and unrelated information. (2) Video Case Study: As illustrated in Figure~\ref{fig:video_case_study}, due to the detecting and repairing ability of our HumanCalibrator, we can repair the first and last frames of a video and regenerate the intermediate frames with a keyframe interpolation model. It shows that with the repaired first and last frames and the original prompt, our Human Calibrator can repair abnormalities while maintaining the other content of the video. (3) Generalization-level: Our HumanCalibrator also demonstrates strong performance on outputs generated by other generative models, with the results shown in Figure~\ref{fig:other_models}. More cases are shown in the Appendix~\ref{sec:more_cases}.

%% file: sec/7_conclusion.tex
\section{Conclusion}
\label{sec:conclusion}
In this paper, we propose HumanCalibrator, a fine-grained level abnormal detection and repair framework in AIGC visual content with two datasets across different domains. It can detect abnormal body parts, and repair the abnormality while maintaining the other visual content. However, there are still limitations to our proposed framework, e.g. the predefined abnormal human body class limits the generalizability. In the future, we plan to extend our method to support more visual categories and more types of abnormalities.

%% file: sec/X_suppl.tex
\clearpage
\setcounter{page}{1}
\maketitlesupplementary

\setcounter{section}{0}
\setcounter{figure}{0}

\renewcommand\thesection{\Alph{section}}
\renewcommand\thefigure{S\arabic{figure}}
\renewcommand\thetable{S\arabic{table}}

\textbf{We highly recommend watching the supplementary video, as it comprehensively demonstrates our proposed task and the results of our proposed HumanCalibrator.}

\textcolor{red}{Disclaimer: The supplementary material includes images that may be unsettling or discomforting to some readers. We have removed all personal information from the cases and applied mosaic to some images that may cause discomfort.}

\section{Details in HumanCalibrator}
\label{sec:detials_in_human_crafter}
\subsection{Model Usage}
In addition to using LLaVAv1.5-7B as the base model for the Absent Human-body Detector, the other models employed in the HumanCalibrator are as follows: (1) The inpainting model $R$, which utilizes StableDiffusion2-Inpainting\footnote{https://huggingface.co/stabilityai/stable-diffusion-2-inpainting, Rombach, R. et al. (2022). High-Resolution Image Synthesis With Latent Diffusion Models. In Proc. CVPR2022.}, (2) the grounding model $G$, which adopts GroundingDINO\footnote{https://github.com/IDEA-Research/GroundingDINO, Liu, S. et al. (2023). Grounding dino: Marrying dino with grounded pre-training for open-set object detection. arXiv}, and (3) the video interpolation model, which employs CogVideoX-Interpolation\footnote{https://huggingface.co/feizhengcong/CogvideoX-Interpolation} based on CogVideoX~\cite{yang2024cogvideox}.

\subsection{Other Implementation Details}
Additional details in our HumanCalibrator are as follows: (1) To improve the repair quality of the overall human photo, we expand the bounding box of the abnormal region before applying inpainting. This ensures better visual quality between the inpainted and surrounding regions. (2) Since the inpainting model inevitably leads to a decline in overall image quality, we apply 2$\times$ super-resolution processing to the inpainted images. It is worth noting that, for a fair comparison, no super-resolution processing is applied in any of the comparisons in Table~\ref{tab:human_and_consistence}. (3) To better adapt the Absent Human-body Detector, trained on real-world COCO datasets, for application in AIGC, we perform semantic detection on each absent region identified by the detector using the Grounding Model $G$. If the same semantic content is detected, the result from this iteration of the Absent Human-body Detector is discarded.

\section{Details of Baselines and Analysis}
\label{sec:detial_baselines}
\subsection{Baseline for COCO Human-Aware Val}
The COCO Human-Aware Val dataset only contains absent abnormalities caused by masking out body parts. Since it is derived from real-world images and includes only the ``absent'' category of abnormalities, our comparisons on this dataset primarily focus on two objectives: (1) demonstrating the deficiency of existing VLMs in abnormality perception and (2) evaluating the performance of our trained Absent Human-body Detector (AHD). 

Evaluating the baseline of CLIP on COCO Human-Aware Val: Similar to other methods, we transform the different types of abnormalities into a classification problem. The CLIP model selects the text with the highest matching score to the image as its predicted answer. The specific text categories are as follows:
\begin{itemize}
    \item ``\textit{The person in the picture has absent head.}''
    \item ``\textit{The person in the picture has absent ear.}''
    \item ``\textit{The person in the picture has absent arm.}''
    \item ``\textit{The person in the picture has absent hand.}''
    \item ``\textit{The person in the picture has absent foot.}''
    \item ``\textit{The person in the picture has absent leg.}''
    \item ``\textit{The person in the image has no abnormalities.}''
\end{itemize}

Evaluating the baseline of Generative VLMs, we use the following prompt to like VQA tasks~\cite{liu2023llava, chen2024far} to prompt the VLMs:
\begin{itemize}
    \item ``\textit{Are there any absent body parts in the person shown in the image? If yes, please answer from `head', `arm', `leg', `foot', `hand', or `ear'; otherwise, please answer `no'. Answer the question using a single word:}''
\end{itemize}

\subsection{Baseline for AIGC Human-Aware 1K}
\begin{figure}[!t]
  \centering
   \includegraphics[width=1\linewidth]{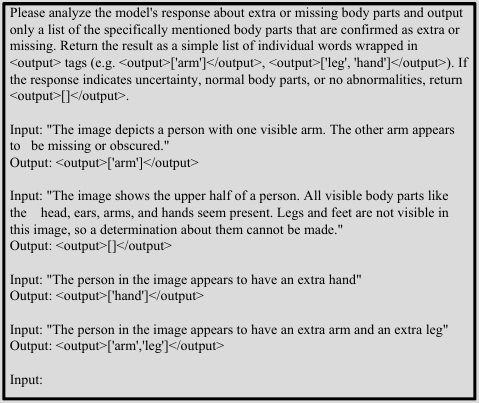}
   \vspace{-20pt}
   \caption{Prompt for post-processing the VLM output.}
   \label{fig:post_trans_prompts}
   \vspace{-10pt}
\end{figure}

Unlike the COCO Human-Aware Val dataset, the AIGC Human-Aware 1K dataset includes all categories of abnormalities. For CLIP, we directly add additional abnormal categories and use a similar classification approach to evaluate its performance. For Generative VLMs, we adopt a simpler method tailored to VLMs. Specifically, we separately ask whether there were abnormalities in the ``redundant'' category and the ``absent'' category. Additionally, since the abnormalities in AIGC Human-Aware 1K are diverse in number, we do not constrain the model's responses to a single word, i.e., we do not use ``answer the question using a single word''. After receiving the responses, we use an LLM for post-processing to produce formatted data suitable for accuracy calculation. Since these baseline VLMs perform weakly on FHAD, we try various prompts per model to optimize performance in our experiments. The prompts that yielded the best performance are displayed below:
\begin{itemize}
    \item For \textbf{LLaVA-34B}:
    \begin{itemize}
        \item In ``Absent Abnormality Detection'': ``\textit{Are there any missing body parts in the person shown in the image? If so, please answer the precise part:}''
        \item In ``Redundant Abnormality Detection'': ``\textit{Are there any extra body parts in the person shown in the image? If so, please answer the precise part:}''
    \end{itemize}
    
    \item For \textbf{Intern VL2-26B}:
    \begin{itemize}
        \item In ``Absent Abnormality Detection'': ``\textit{According to the human anatomical structure, are there any missing body parts in the person shown in the image? If so, please answer the precise part:}''
        \item In ``Redundant Abnormality Detection'': ``\textit{According to the human anatomical structure, are there any extra body parts in the person shown in the image? If so, please answer the precise part:}''
    \end{itemize}

    \item For \textbf{GPT-4o}:
    \begin{itemize}
        \item In ``Absent Abnormality Detection'': ``\textit{It is a common sense that all human being has one head, two ears, two hands, two arms, two legs and two foots, are there any missing body parts which I discussed in the person shown in the image? If so, please answer the precise part:}''
        \item In ``Redundant Abnormality Detection'': ``\textit{It is a common sense that all human being has one head, two ears, two hands, two arms, two legs and two foots, are there any extra body parts which I discussed in the person shown in the image? If so, please answer the precise part:}''
    \end{itemize}
\end{itemize}
For the post process for the response of VLMs (Note that, we use the GPT4o-mini to post-process the response) as shown in Figure~\ref{fig:post_trans_prompts}.

\subsection{Baseline Analysis}
\begin{figure*}[!t]
  \centering
   \includegraphics[width=1\linewidth]{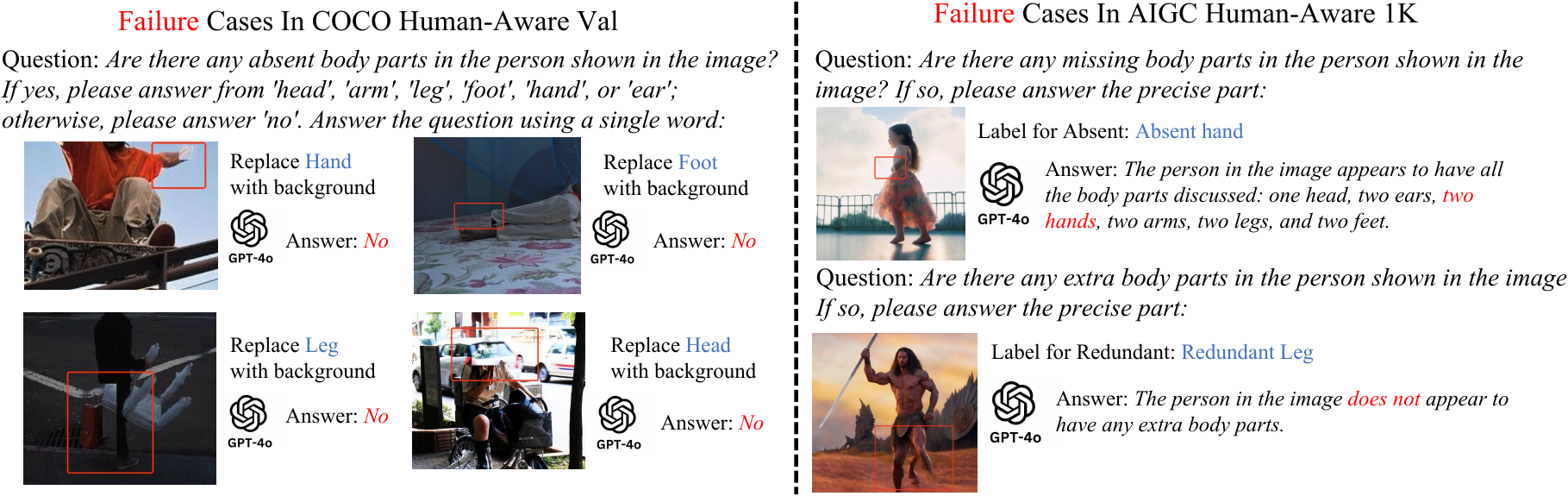}
   \vspace{-20pt}
   \caption{Failure cases of the powerful VLM (GPT-4o) on COCO Human-Aware Val and AIGC Human-Aware 1K. For COCO Human-Aware, it is observed that despite generating distinctly anomalous images, GPT-4o still responds with a definitive ``No''. In the case of AIGC Human-Aware 1K, even though GPT-4o is aware of the components that constitute a normal human body, it fails to recognize or respond to abnormalities. Note that our prompt includes the category of abnormalities, which simplifies the task; however, GPT-4o still struggles to perform effectively, resulting in poor baseline performance.}
   \label{fig:baseline_fail_case}
   \vspace{-10pt}
\end{figure*}

Our work is based on a key assumption: that existing powerful VLMs fail to perform abnormality detection, a task that is exceptionally simple for humans. We provide a detailed case analysis of their poor performance. Specifically, there are two primary reasons for this under-performance: (1) a lack of understanding of human body structure, and (2) a misinterpretation of abnormalities. We present examples from \textbf{real test} in Figure~\ref{fig:baseline_fail_case}.

\subsection{Pose Condition}
Since the code for HumanRefiner~\cite{fang2024humanrefiner} is unavailable and our objective differs fundamentally, we only reproduce its step of using pose as an additional constraint to ensure no abnormalities in the number of body parts. Specifically, for the input human photo, we use MMPose\footnote{MMPose Contributors. (2020). OpenMMLab Pose Estimation Toolbox and Benchmark. Retrieved from https://github.com/open-mmlab/mmpose} to extract the human pose and then employ Stable-Diffusion-v1.5\footnote{https://huggingface.co/stable-diffusion-v1-5/stable-diffusion-v1-5, Rombach, R. et al. (2022). High-Resolution Image Synthesis With Latent Diffusion Models. In Proc. CVPR2022.} with t2iadapter\_keypose\footnote{https://github.com/TencentARC/T2I-Adapter, Mou, C. et al. (2023). T2i-adapter: Learning adapters to dig out more controllable ability for text-to-image diffusion models. arXiv} as a pose-conditioned method to regenerate the entire image.

\section{Why do current VLMs lack the ability to perceive abnormality?}
\label{sec:why_lack}
Our extensive experiments demonstrate that existing VLMs are unable to perceive human abnormalities (some cases are shown in Figure~\ref{fig:baseline_fail_case}), even though this task is very simple for humans, and both we humans and the models are trained on a large amount of normal data. We believe that the drawbacks arise from the simplistic image-text alignment approach of existing VLMs, which lacks perception of content and, consequently, an understanding of human body structure. Additionally, the existing VLMs underutilize the data and are undertrained, and the proportion of human subjects in the training data may not be substantial. In our work, we utilize the correlation among human body structures to train our absent human-body detector.

\section{AIGC Human-Aware 1K Annotation}
\label{sec:data_annotation}
\begin{figure*}[!t]
  \centering
   \includegraphics[width=1\linewidth]{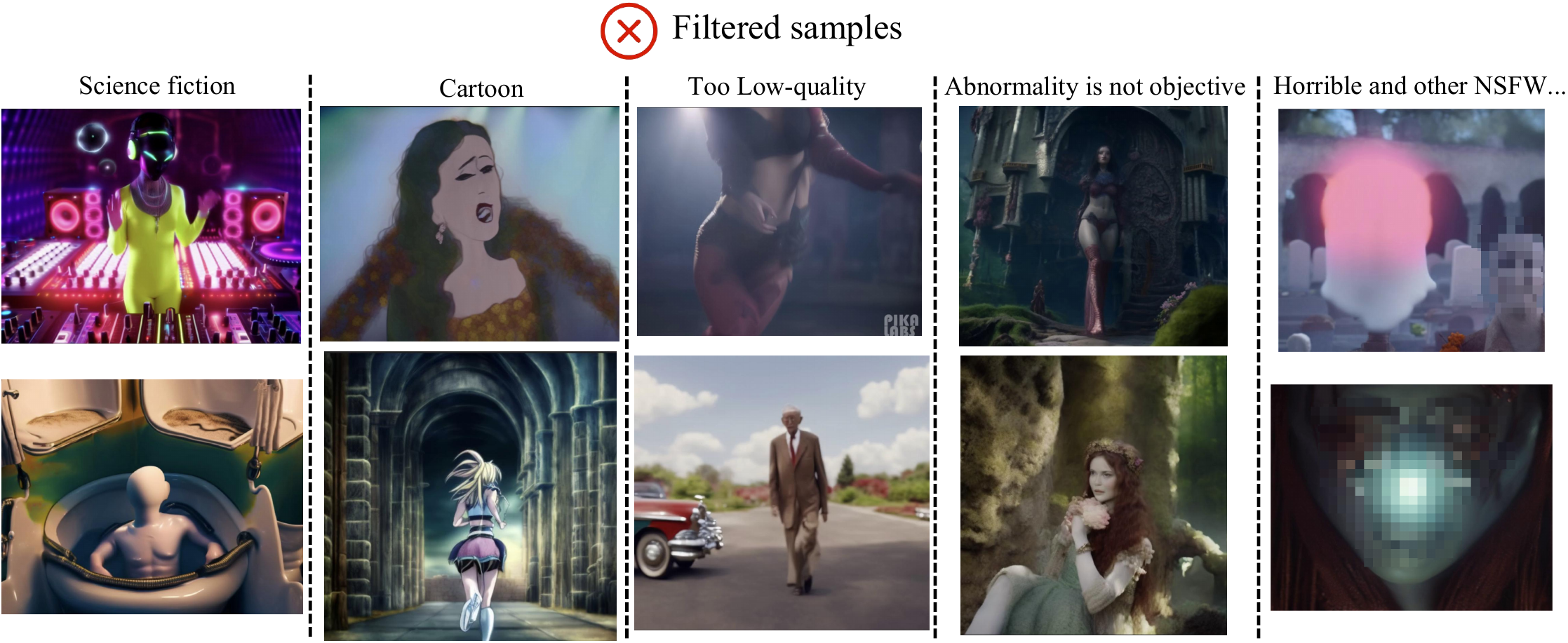}
   \vspace{-20pt}
   \caption{Categories and examples filtered out during the annotation process for AIGC Human-Aware 1K. The goal of our proposed task, ``Fine-grained Human-body Abnormality Detection'', is to determine whether the body structure in a given human photo could exist in the real world. Thus our annotated data are grounded in real-world contexts, leading us to exclude images of genres such as science fiction and cartoons. Additionally, to enhance the dataset's quality, we filter out samples where the specific abnormality cannot be ascertained or where the abnormality is controversial, labeled as ``Too Low-quality'' or ``Abnormality is not objective''. All NSFW images have also been excluded, and \textbf{\textit{the displayed samples have been processed with mosaic}}. These rigorous criteria not only ensure the quality of our AIGC Human-Aware 1K dataset but also explain why annotating a large number of data for the training process directly from AIGC is costly.}
   \label{fig:filtered_images}
   \vspace{-10pt}
\end{figure*}

The target of our proposed task, ``Fine-grained Human Abnormality Detection'', is to detect whether the human photos in AIGC exhibit abnormalities that render them impossible to exist in the real world. This imposes two requirements on our evaluation data: (1) The annotated abnormalities must be objective, avoiding controversial cases caused by ambiguity or other factors. (2) The human photos in the annotations must appear in real-world environments, which necessitates selecting realistic styles for annotation and excluding sci-fi or cartoon-style images. In Figure~\ref{fig:filtered_images}, we demonstrate examples of cases that are manually filtered out during the annotation process. After the initial annotation, to ensure data objectivity, we conduct multi-round and multi-reviewer checks on the data labels, removing any remaining controversial annotations. This process ensures the quality of our proposed AIGC Human-Aware 1K dataset. We provide statistics on the number of different annotation types in AIGC Human-Aware 1K, as shown in Table~\ref{tab:aigc_statisitc}.

\begin{table}[!b]
  \centering
\resizebox{1\columnwidth}{!}{
    \begin{tabular}{cccc}
    \toprule
    Type & Absent& Redundant& No Abnormality \\
    \midrule
    Number & 649   & 158   & 343 \\
    \bottomrule
    \end{tabular}%
    \vspace{-5pt}
    }
    \caption{Statistics on the Number of Annotation Types in AIGC Human-Aware 1K}
  \label{tab:aigc_statisitc}%
  \vspace{-5pt}
\end{table}%

\section{Metric Details}
\label{sec:detial_of_metrics}
It is essential to emphasize that a comprehensive evaluation of our proposed task requires the integration of multiple metrics. Specifically, we employ Accuracy (ACC) and False Discovery Rate (FDR) as detection metrics to ascertain the correct identification of existing abnormalities. Furthermore, we utilize perceptual metrics to evaluate the reasonableness of the identified abnormal locations and to assess the quality of the repairs to these abnormalities. Additionally, we use the Fréchet Inception Distance (FID) and Latent Consistency to examine the similarity between our repaired human photos and the original human photos, demonstrating the granularity of our repairs; i.e, we only repair the abnormal areas while preserving the other content.

\subsection{Human CLIP Scores}
Since our task focuses on repairing the human body in a given human photo, directly using the original prompt to calculate the CLIP score is not ideal, as it includes substantial background and camera-related information. Instead, we utilize GPT4o-mini to extract prompts specifically related to the human body to evaluate whether our repair improves the correlation with human-related prompts, a metric we refer to as the Human CLIP Score. An example case is shown below:
\begin{itemize}
    \item Original Prompt: ``\textit{A girl with long hair is walking on the avenue in the forest, with a gentle breeze blowing her hair and falling leaves fluttering in the wind. The girl looks melancholy in the distance.}''
    \item Processed Human Prompt: ``\textit{A girl with long hair is walking on the avenue in the forest, looking melancholy into the distance.}''
\end{itemize}

\subsection{Human Concept Scores}
Compared to the Human CLIP Score, which focuses more on the quality of the human body in the repaired human photo, the Human Concept Score emphasizes evaluating whether the repaired human conceptually aligns more closely with the distribution of ``human'' as understood by CLIP, trained on extensive real-world data. To verify this, we use a straightforward method: calculating the similarity between the human photo before and after repair and the prompt ``an image contains human'' to examine whether the repaired human better matches CLIP's concept of a human existing in the real-world which learned from diverse real-world training data.

\subsection{Visual Consistency}
For the FID, we treat the repaired images as generated images and calculate the distributional discrepancy between them and the original images. For Latent Consistency, we encode the images into the latent space via the CLIP Visual Encoder and compute the cosine similarity between the original and repaired images.

\section{Cases in COCO Human-Aware Val}
\label{sec:coco_val_case}
\begin{figure}[!t]
  \centering
   \includegraphics[width=1\linewidth]{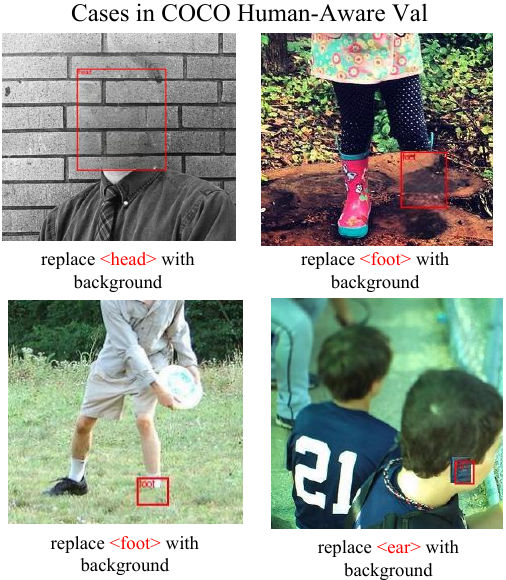}
   \vspace{-20pt}
   \caption{Examples of the AHD on COCO Human-Aware Val. The \textcolor{red}{red boxes} indicate the predictions made by AHD. It is observable that AHD, trained utilizing the correlation within human body structures, can accurately identify the location and type of artificially created abnormalities. Note that all personal information has been removed from the cases displayed. The training set created from the COCO Train Split is in a similar format.}
   \label{fig:cases_in_coco_val}
   \vspace{-10pt}
\end{figure}

We also provide examples of the Absent Human-body Detector's performance on the COCO Human-Aware Val, as shown in Figure~\ref{fig:cases_in_coco_val}. It shows that the trained Absent Human-body Detector accurately identifies the locations and the type of artificially created abnormalities.

\section{More Cases}
\label{sec:more_cases}
\begin{figure*}[!t]
  \centering
   \includegraphics[width=1\linewidth]{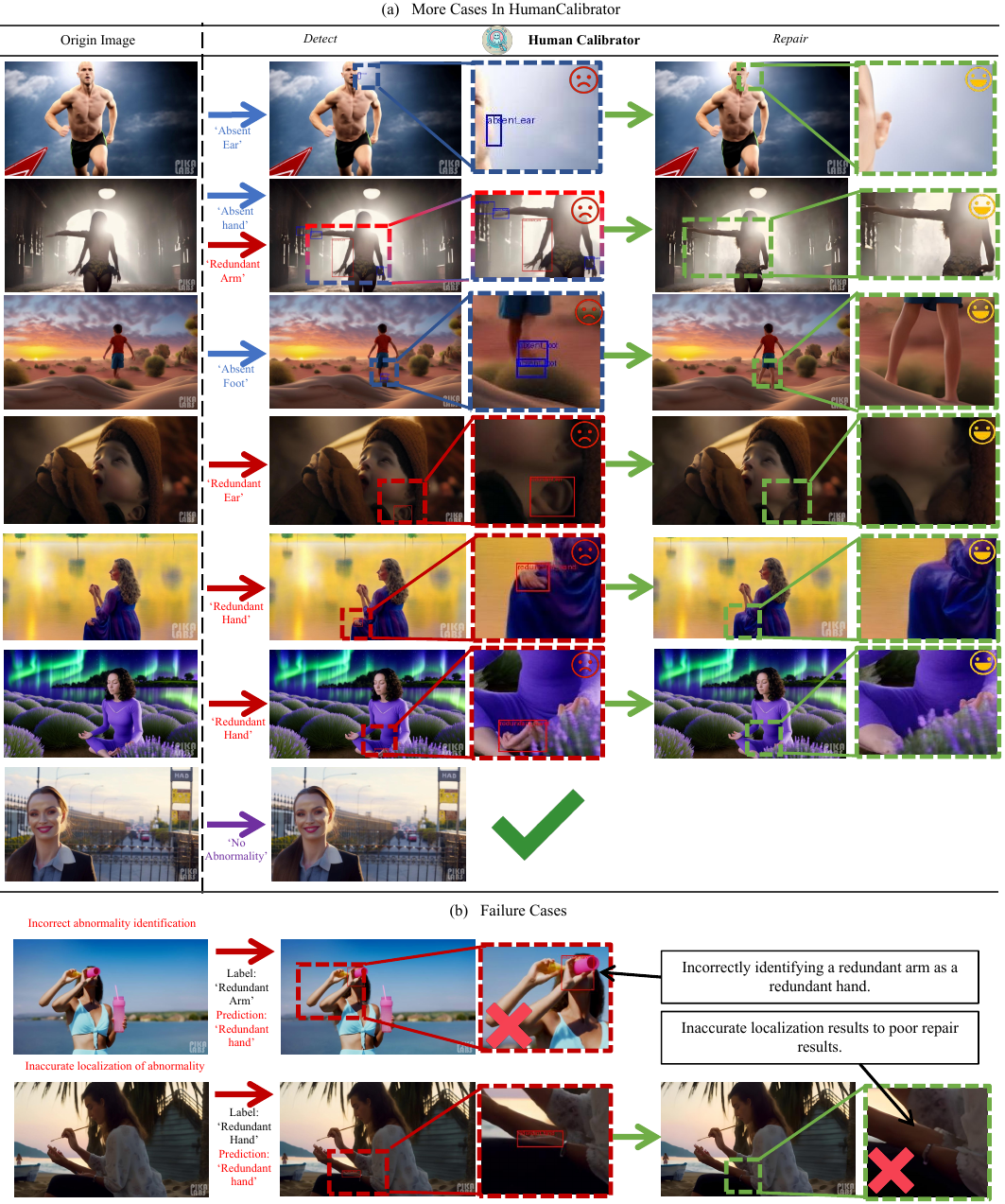}
   \vspace{-20pt}
   \caption{More Cases in HumanCalibrator and some failure cases.}
   \label{fig:more_cases}
   \vspace{-10pt}
\end{figure*}

In Figure~\ref{fig:more_cases} (a), we provide additional examples, including results for test cases with no abnormalities or in complex scenarios. Additionally, we present several failure cases, which primarily fall into two categories: the first involves incorrect abnormality identification by the HumanCalibrator, and the second involves inaccurate localization of abnormalities, leading to reduced repair quality. These are illustrated in Figure~\ref{fig:more_cases} (b).

